\documentclass[letterpaper, 10 pt, conference]{ieeeconf}  

\usepackage{times}

\usepackage{graphics} 
\usepackage{epsfig} 
\usepackage{amsmath} 
\usepackage{amssymb}  
\usepackage{graphicx}
\makeatletter
\let\NAT@parse\undefined
\makeatother
\usepackage[colorlinks=true, linkcolor=blue, urlcolor=blue, citecolor=red, breaklinks=true]{hyperref}
\usepackage{cleveref}
\usepackage{csquotes}

\title{\LARGE \bf
The SLAM Confidence Trap
}

\author{Sebastian Sansoni$^{1}$, Santiago Ramón Tosetti$^{1}$
\thanks{The authors are with the Instituto de Automática (INAUT), Universidad Nacional de San Juan (UNSJ-CONICET), San Juan, Argentina.
        {\tt\small \{ssansoni, stosetti\}@inaut.unsj.edu.ar}}%
}

\usepackage{tikz}

\newcommand{\mycopyright}{
\begin{tikzpicture}[remember picture, overlay]
    \node[anchor=north, yshift=-0.5cm] at (current page.north) {
        \parbox{\textwidth}{
            \centering
            \scriptsize 
            \textcolor{black}{Accepted to the 2025 IEEE International Conference on Advanced Robotics (ICAR), DOI: \href{https://doi.org/10.1109/ICAR65334.2025.11338690}{10.1109/ICAR65334.2025.11338690}} \\
            \vspace{2pt}
            \textcolor{black}{\copyright 2025 IEEE. Personal use of this material is permitted. Permission from IEEE must be obtained for all other uses, in any current or future media, including reprinting/republishing this material for advertising or promotional purposes, creating new collective works, for resale or redistribution to servers or lists, or reuse of any copyrighted component of this work in other works.}
        }
    };
\end{tikzpicture}
}

\begin{document}

\maketitle
\mycopyright 

\begin{abstract}

The SLAM community has fallen into a \enquote{Confidence Trap} by prioritizing benchmark scores over principled uncertainty estimation. This yields systems that are geometrically accurate but probabilitistically inconsistent and brittle. We advocate for a paradigm shift where the consistent, real-time computation of uncertainty becomes a primary metric of success.

\end{abstract}

\section{INTRODUCTION}

The integrity of autonomous navigation faces a significant and growing threat from Global Navigation Satellite System (GNSS) spoofing. The wide availability of low-cost platforms, such as software-defined radios (SDRs), enables malicious attacks where falsified signals systematically deceive navigation systems, thus compromising the reliability of GNSS-based positioning and timing \cite{radovs2024recent}. These are not mere interferences; they are sophisticated attacks that can cause an aircraft's Inertial Reference System (IRS) to diverge significantly as its position is continuously \enquote{corrected} with malicious data. The European Union Aviation Safety Agency (EASA) has issued urgent safety bulletins describing these events, which create \enquote{significant safety risks} by leading aircraft dangerously off-course, sometimes towards hostile airspace \cite{easa_sib_2022}. Spoofing is even a suspected contributing factor in ongoing aviation investigations, such as the 2024 Azerbaijan Airlines crash \cite{flightradar24_azerbaijan_crash}.

This dramatic real-world failure, however, is not an isolated problem for aviation. It is the most visible symptom of a deeper, more systemic issue we term \textit{the SLAM Confidence Trap}: the dangerous gap between a system's performance on offline benchmarks and its robustness in the real world. A navigation system under attack perfectly illustrates this trap. The system might be capable of achieving near-perfect trajectory accuracy on a standard dataset, yet it becomes catastrophically brittle when a core assumption, the trustworthiness of its primary sensor, is violated. This highlights a fundamental question: how can a system know when its localization source is not just noisy, but actively lying? The challenge is not just detecting spoofing, but having a robust framework to reject the deceptive data and continue operating safely.

This vulnerability has been inadvertently cultivated by the academic SLAM community’s own evaluation standards. For over a decade, progress has been predominantly measured by the reduction of Absolute Trajectory Error (ATE) and Relative Pose Error (RPE) on a set of standardized, static benchmarks \cite{sturm2012benchmark,geiger2012we}. While invaluable for development, this benchmark-centric culture has inadvertently promoted a \enquote{collect-then-process} paradigm. In this approach, systems acquire vast amounts of data with quality only being assessed after the collection is complete, often leading to models with significant, unquantified uncertainty and no mechanism for real-time corrective action \cite{stachniss2005information, sansoni2025optimizing}. Consequently, the crucial aspect of the original probabilistic SLAM formulation, the explicit and consistent estimation of uncertainty, has been neglected.

This focus on retrospective accuracy creates a dangerous evaluation gap. A system may achieve centimeter-level accuracy on a dataset but have no way to assess the reliability of its estimate when deployed in a new, unstructured, or even adversarial environment. When an external sensor is compromised, the system becomes brittle. More insidiously, when the SLAM system itself makes an error, for instance a false loop closure, it can become overconfident in its incorrect state. This represents a failure of internal consistency that is often undetectable by standard metrics.

This paper argues that escaping the \enquote{SLAM Confidence Trap} requires a fundamental paradigm shift. We must move from evaluating systems solely on benchmark accuracy to prioritizing their real-time, internal uncertainty-awareness. We contend that a system’s ability to know what it does not know is as critical as its ability to produce an accurate estimate. To this end, our contributions are:

\begin{enumerate}
\item A critical analysis of the evolution of SLAM, highlighting the trend away from explicit uncertainty estimation in modern optimization-based methods.
\item A proposal to shift the focus of SLAM evaluation from purely external, offline error metrics to internal, real-time metrics of system confidence and consistency.
\end{enumerate}

By making uncertainty a first-class citizen in SLAM design and evaluation, we can build the robust, resilient, and safe autonomous systems that real-world applications demand.

\section{THE STATE OF THE ART: A TALE OF LOST COVARIANCE}

The history of SLAM can be narrated as a tale of trade-offs, where the pursuit of accuracy and scalability led to a subtle but significant shift away from explicit probabilistic modeling. This evolution occurred across distinct eras, each contributing to the current \enquote{evaluation gap}.

\subsection{The Probabilistic Era (Pre-2010s)}

Early SLAM solutions were fundamentally probabilistic, an approach implemented through the Extended Kalman Filter (EKF) in the pioneering work of Smith, Self, and Cheeseman \cite{smith1990estimating}. Their key insight was to formulate SLAM as a single estimation problem by representing the robot's pose and all landmark locations within one augmented state vector. In this formulation, a full covariance matrix captured all uncertainties and, crucially, their cross-correlations, providing the blueprint for the \enquote{stochastic map}.


This theoretical elegance, however, was initially met with skepticism, as it was widely assumed that the map error would execute a random walk with unbounded error growth. This foundational uncertainty was resolved by Dissanayake et al. \cite{dissanayake2001solution}, who delivered the formal proof that the total map uncertainty decreases monotonically and converges to a stable, bounded value, establishing the theoretical consistency of EKF-SLAM. However, this guarantee of consistency applies to the ideal linear case. The work of Castellanos et al. \cite{castellanos2004limits} later demonstrated that in real-world, non-linear scenarios, the EKF's reliance on linearization introduces systematic errors. This leads to a critical failure of consistency, where the filter becomes overly optimistic, with its calculated covariance becoming decoupled from the true estimation error \cite{castellanos2004limits}.
The very foundation of probabilistic SLAM, a consistent covariance, was thus shown to be brittle in practice.
 
Alongside the EKF, a parallel approach emerged using Rao-Blackwellized Particle Filters (RBPF), pioneered conceptually by Murphy \cite{murphy1999bayesian}. This non-parametric method represents the posterior over robot trajectories with a weighted particle cloud, where each particle carries an individual map. System uncertainty is therefore not captured by an explicit covariance matrix, but is implicitly encoded in the diversity of the particle set. The primary threat to consistency for this method is particle depletion, a phenomenon where repeated resampling causes the filter to collapse to a single hypothesis, a catastrophic loss of diversity equivalent to the EKF’s over-confident state. A key advancement to mitigate this was the improved proposal distribution introduced in FastSLAM 2.0 \cite{montemerlo2003fastslam}, a principle later refined and paired with the adaptive resampling of GMapping \cite{grisetti2007improved}, both designed to preserve a consistent uncertainty representation. Thus, while particle filters could represent multi-modal distributions and avoid linearization errors, their own version of the \textit{confidence trap} required sophisticated sampling strategies to ensure a consistent representation of uncertainty. Ultimately, the forward-only nature of both filtering approaches, unable to retrospectively correct past states, represented a fundamental limitation in achieving globally consistent uncertainty estimates and paved the way for the next generation of SLAM algorithms.

\subsection{The Optimization Revolution: A Paradigm Shift}

In response to the inherent limitations of filtering approaches, a paradigm shift toward batch optimization emerged, a revolution conceptually sparked by seminal works like that of Lu and Milios \cite{lu1997globally}. Their core insight was to re-frame SLAM not as a recursive state estimation problem, but as a large-scale geometric graph to be optimized. By representing robot poses as nodes and relative transformations from odometry and scan-matching as constraints, their method could find a globally consistent solution for the entire trajectory at once. This was a powerful departure from the forward-only nature of filters, offering a solution to the critical problem of closing large loops and achieving global consistency.

However, this foundational work also planted the seed of the \enquote{SLAM Confidence Trap}. The elegance of the optimization relied on weighting each constraint by its covariance, but the authors themselves conceded that the assumptions for deriving the system's covariance matrices  were \enquote{probably difficult to justify}, opting instead for heuristics that were \enquote{reasonable ones in practice} \cite{lu1997globally}. This pragmatic choice marked a pivotal shift where probabilistic rigor was subtly traded for geometric accuracy. This is starkly reflected in the paper's evaluation, which moved away from visualizing uncertainty ellipses, a hallmark of the EKF era, to plotting pose error against ground truth. The primary question of SLAM research began to drift from \enquote{How consistent is my probabilistic belief?} to \enquote{How low is my final trajectory error?}, setting a precedent that would influence the field for decades.

\begin{figure*}[t]
    \centering
    \includegraphics[height=0.3\textheight]{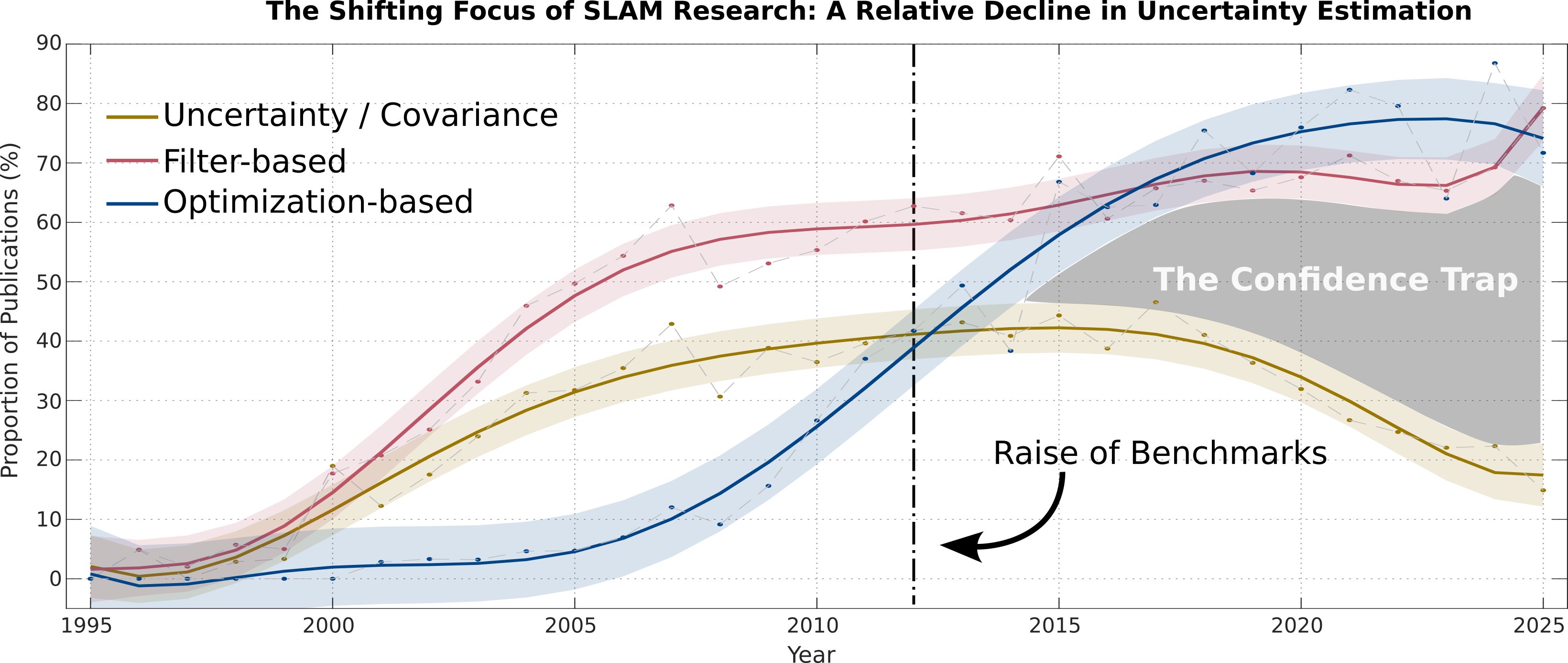}
    \caption{An illustration of the historical shift in SLAM research focus, depicting the relative decline in explicit uncertainty estimation. The graph plots the proportion of publications over time that mention key terms related to: (i) filter-based methods (e.g., EKF, particle filters), (ii) optimization-based methods (e.g., graph optimization, bundle adjustment), and (iii) uncertainty or covariance. A clear trend emerges post-2010, coinciding with the \enquote{Raise of Benchmarks}, where optimization-based approaches dominate while the relative focus on uncertainty estimation diminishes. This divergence highlights the emergence of  \enquote{the Confidence Trap}, the growing gap between geometric accuracy and probabilistic consistency. 
    The data and code used to generate this figure are publicly available at \href{https://github.com/Seba-san/SLAM-confidence-bibliometric}{https://github.com/Seba-san/SLAM-confidence-bibliometric}}.
    \label{fig:1}
 \end{figure*}

This pragmatic compromise, however, did not go unchallenged. The critical task of reconciling the global consistency of graph optimization with the demands of real-time operation was met by the breakthrough of Incremental Smoothing and Mapping (iSAM) \cite{kaess2008incremental}. This framework proved that it was possible to maintain the full state trajectory, the key to probabilistic consistency, while operating efficiently online. By working with a factorized representation of the system's information matrix, iSAM's pivotal innovation was a method to recover \textit{exact} marginal covariances on demand. This made robust, uncertainty-aware decisions like data association feasible in real time, offering a clear blueprint where performance did not require sacrificing probabilistic rigor.

This optimization-first philosophy was soon encapsulated and massively amplified by the release of g²o, an open-source C++ framework that democratized graph-based optimization for the entire field \cite{kummerle2011g2o}. The framework's design is itself the ultimate evidence of the trap: its core interface requires the user to define a geometric error function and provide a corresponding information matrix ($\Omega$), but it is entirely agnostic to the probabilistic provenance or consistency of that matrix. This architectural choice effectively standardized the use of heuristic weights, making it trivial to prioritize a low geometric error while overlooking the validity of the underlying probabilistic model. Because this powerful and accessible tool removed the implementation barrier, the community's metrics of success shifted definitively. Uncertainty ellipses vanished from result plots, and formal validation methods like the Chi-squared ($\chi^2$) test, a well-established tool for verifying the statistical consistency of the solution \cite{grisetti2011tutorial} and actively used in state-of-the-art research of the era \cite{carlone2014fast}, became a rarity. The new measures of success were computational speed and final trajectory error, cementing a paradigm where geometric accuracy reigned supreme.

This systemic pivot towards geometric accuracy at the expense of probabilistic rigor is quantitatively illustrated in Fig. \ref{fig:1}. The figure tracks the proportion of publications mentioning different SLAM paradigms, highlighting the dramatic rise of optimization-based methods that coincides with the release of influential benchmarks, and the corresponding relative decline in research explicitly addressing uncertainty. This trend is clearly reflected in the academic literature that followed.  Comprehensive surveys of the period, such as the one by Taketomi et al., analyze and categorize the field's progress between 2010 and 2016 based on architectural choices like feature-based versus direct methods, or the presence of global optimization and loop closure \cite{taketomi2017visual}. In these analyses, which represent the consensus view of the field's evolution, the explicit estimation and validation of uncertainty are seldom treated as a core evaluation criterion. The academic conversation itself had shifted, cementing a new definition of \enquote{state-of-the-art} where benchmark performance overshadowed probabilistic rigor.


The paradigm shifted definitively in 2012 with the release of the two most influential benchmarks in visual SLAM. The TUM RGB-D benchmark for indoor robotics \cite{sturm2012benchmark} and the KITTI Vision Benchmark Suite for autonomous driving \cite{geiger2012we} established Absolute Trajectory Error (ATE) and Relative Pose Error (RPE) as the universal metrics of success. While invaluable for standardizing comparisons, this new evaluation framework defined a system's quality almost exclusively by its retrospective geometric accuracy, effectively decoupling the notion of "state-of-the-art" from the requirement of maintaining and validating a consistent probabilistic belief.

This growing disconnect between benchmark performance and real-world resilience did not go unnoticed. In a landmark 2016 position paper, a consortium of the field's leading researchers effectively declared the end of the geometric accuracy era and called for a new \enquote{Robust-Perception Age} \cite{cadena2016past}. Their definition of future progress was not based on lower trajectory errors, but on qualities like fail-safe mechanisms, high-level scene understanding, and task-driven perception. Most critically, they delivered a stark verdict on the state-of-the-art systems of the time, stating that despite their benchmark success, \enquote{None of the existing SLAM approaches provides these capabilities} for robust, failure-aware operation. This declaration from the very architects of modern SLAM represents the most formal acknowledgment of the \enquote{SLAM Confidence Trap}, highlighting that even systems achieving unprecedented accuracy were fundamentally brittle and lacked the crucial ability of self-assessment.

The benchmark-centric paradigm, cemented by 2012, has largely persisted into the present day. A comprehensive 2022 survey by Kazerouni et al. confirms this inertia, showing that established SLAM systems are still primarily judged by their geometric accuracy on datasets like KITTI \cite{kazerouni2022survey}. However, the same survey unwittingly chronicles the beginning of a fundamental disruption. It details the rise of Deep Learning, a technology that, while initially used to simply improve scores on the same benchmarks, is now forcing a paradigm shift from within. The advent of unsupervised and semantic approaches has created a practical necessity for systems to estimate their own uncertainty to be robust, and it has introduced new performance metrics that go beyond mere trajectory error. This shift, where uncertainty is reborn not from a return to probabilistic theory but from the pragmatic demands of a new technology, marks the dawn of a new era.

\subsection{The Machine Learning Renaissance}

More recently, the field of 3D representation has been revolutionized by machine learning, particularly through methods like Neural Radiance Fields (NeRF) \cite{mildenhall2020nerf} and 3D Gaussian Splatting \cite{kerbl2023gaussian}. These techniques can generate photorealistic novel views of a scene from a collection of images, learning a continuous or semi-explicit representation of scene geometry and appearance. Their results are visually stunning and have pushed the state of the art in scene representation.

However, from the perspective of uncertainty, these methods largely perpetuate the paradigm of external evaluation. In their original formulations, they do not inherently provide a measure of geometric or photometric uncertainty for their reconstructions. Quality is assessed by comparing rendered images to a held-out test set using image-space metrics like PSNR (Peak Signal-to-Noise Ratio) or SSIM (Structural Similarity Index Measure). While powerful for visualization, they do not offer a mechanism for the system to self-assess its confidence in unobserved parts of the scene or to quantify the reliability of its learned geometry, a critical component for robotic interaction and decision-making. 

\subsection{The Resulting Evaluation Gap}

The historical focus on offline geometric accuracy has created a dangerous evaluation gap, a reality starkly revealed when state-of-the-art SLAM systems are deployed in real-world environments like precision agriculture. These settings defy the assumptions of the structured, static worlds found in canonical benchmarks. Instead, they present pervasive environmental dynamics, such as wind-blown foliage, and perception-degrading conditions, like the highly repetitive textures of crop rows and severe illumination changes under canopies. These are not the discrete, moving objects handled by systems like DynaSLAM \cite{bescos2018dynaslam}, but a more fundamental challenge to the feature tracking and data association capabilities of any SLAM front-end. This performance collapse, demonstrated in multiple comparative studies \cite{cremona2022experimental, capua2018comparative}, validates the long-held concerns about the need for a \enquote{Robust-Perception Age} \cite{cadena2016past} and proves that even recent efforts to create more demanding benchmarks, like the HILTI SLAM Challenge Dataset \cite{helmberger2022hilti}, do not fully capture the unique failure modes that prevent reliable, long-term autonomy in natural environments.

This demonstrated inadequacy has forced the agricultural robotics community to forge its own path, creating evaluation tools and methodologies tailored to their specific problems. This necessity gave birth to challenging, large-scale datasets captured on real farms, such as the Rosario Agricultural Dataset \cite{pire2019rosario} and the University of Bonn's Agricultural  Dataset \cite{chebrolu2017agricultural}. These are not simply new testbeds for ATE leaderboards; they were explicitly designed to capture the complex and dynamic conditions that cause mainstream algorithms to fail. Consequently, the metric for success in this domain shifts fundamentally from abstract geometric accuracy to tangible task-based utility. As detailed in the comprehensive survey by Fasiolo et al., a system’s quality is judged not by its final trajectory error, but by its ability to deliver a coherent NDVI map for crop health analysis, generate a precise 3D model for canopy volume estimation, or provide a stable localization for autonomous spraying \cite{fasiolo2023towards}. This focus on task completion is the pragmatic and necessary escape from the SLAM Confidence Trap, representing a paradigm shift where robustness and reliability are valued not for their own sake, but as essential enablers for the real-world applications of tomorrow.


\subsection{Emerging Probabilistic and Decentralized Paradigms}

Recognizing the limitations of centralized Non-Linear Least Squares (NLLS) solvers, a new wave of research is revisiting probabilistic inference as a core component of SLAM. Recent work, such as the Hyperion framework \cite{hug2024hyperion}, proposes moving away from traditional NLLS towards distributed and inherently probabilistic methods like Gaussian Belief Propagation (GBP).

In such frameworks, the uncertainty (i.e., the covariance) is not a mere by-product to be computed at the end, but a central element of the message-passing algorithm itself. Hug et al. \cite{hug2024hyperion} argue that explicitly modeling uncertainties allows for advanced capabilities, such as the efficient allocation of computational resources to the parts of the estimation problem that are least certain. This represents a significant shift, re-emphasizing that a deep understanding and manipulation of uncertainty is key to building the next generation of scalable, multi-agent, and efficient SLAM systems. This emerging trend further reinforces the argument that the era of treating SLAM as a purely geometric optimization problem is reaching its limits.

\section{The Case for Uncertainty-Awareness: From Passive Metric to Active Driver}


If the era of geometric optimization is indeed reaching its limits, the path forward demands a fundamental paradigm shift. We must move beyond treating uncertainty estimation as an optional extra and recognize it as the cornerstone of robust autonomy. Escaping the Confidence Trap requires embracing uncertainty not as a passive metric for post-hoc evaluation, but as an active driver for intelligent action.  Its importance is threefold: it enables the use of principled representations, allows for robust fault detection, and is a prerequisite for intelligent action.

Uncertainty  is the engine for proactive decision-making. The goal of Active SLAM and Informative Path Planning (IPP) is to plan paths that optimally reduce system uncertainty, thereby improving map quality and localization accuracy with minimal effort. This is fundamentally impossible if the system does not maintain an explicit representation of its own uncertainty.

The evolution of this idea demonstrates the growing sophistication of uncertainty-aware systems. The traditional approach, based on Classical Frontiers \cite{yamauchi1997frontier}, guides a robot simply to the boundary between explored and unexplored space. While effective for maximizing coverage, this strategy is blind to the quality of the existing map. The next great leap was to use the SLAM system's own uncertainty to guide the robot. Seminal work in this area pioneered planning paths to maximize information gain \cite{stachniss2005information} or directly control camera movements to improve landmark triangulation and reduce depth uncertainty in monocular SLAM \cite{davison2005active}. This philosophy, where uncertainty drives action, is the core of true active perception.

This principle has since matured into a core component of modern robotics, leading to sophisticated \enquote{perception-aware} navigation systems. These systems integrate state uncertainty directly into real-time trajectory planners, ensuring that a robot moves not only efficiently but also in a way that guarantees the quality of its own perception along the path \cite{zhang2018perception}.

More recently, in systems that rely on learning-based perception, this concept has reached a new level of abstraction. The new frontier focuses on reducing the epistemic uncertainty of the onboard models themselves. State-of-the-art frameworks now guide robots to regions where a semantic segmentation model is least confident \cite{ruckin2023tro, ruckin2022iros}. This allows the system to gather data that is maximally informative for re-training, representing a significant evolution from simple exploration to true robotic self-improvement. The ultimate goal is a system that not only knows that it is uncertain, but why. As argued by Qin et al. \cite{qin2024towards}, by explaining and attributing uncertainty to specific causes (e.g., 'high uncertainty due to insufficient overlap'), a robot can make more intelligent, symbolic decisions, such as actively repositioning its sensors to mitigate that specific source of uncertainty.

\section{Modern Tools for Uncertainty Quantification} 

To realize the vision of an uncertainty-driven system, especially as perception pipelines become dominated by machine learning, we must equip our robots with the proper tools for self-assessment. While traditional SLAM has grappled with uncertainty through probabilistic representations like covariance matrices, the deep learning community has developed its own rich set of techniques to address this problem from a different perspective. For safety critical robotic applications, a single deterministic prediction is insufficient. The key is to understand why a model is uncertain, which leads to the crucial distinction between two types of uncertainty established in the machine learning literature: aleatoric and epistemic \cite{hullermeier2021aleatoric}.


Aleatoric uncertainty captures inherent and irreducible data noise, such as the thermal noise in an IMU. This is typically modeled by training a network to predict the parameters of a full probability distribution for its output (e.g., a mean and a variance), rather than just a single point estimate \cite{kendall2017uncertainties}. However, for robust decision making in unknown situations, epistemic uncertainty is arguably more critical. It captures the model's own ignorance due to insufficient training data. This uncertainty is reducible and is essential for detecting out of distribution (OOD) samples, allowing a model to effectively report: \enquote{I do not know because I have not seen this before}.


To estimate this critical epistemic uncertainty, several practical techniques have been established. Among the most widespread are \textit{Monte Carlo (MC) Dropout} \cite{gal2016dropout} and \textit{Deep Ensembles} \cite{lakshminarayanan2017simple}. Both methods generate a distribution of predictions for a given input, where the variance serves as a direct measure of epistemic uncertainty. While MC Dropout offers an efficient approximation, Deep Ensembles often achieve state-of-the-art results in both accuracy and uncertainty calibration, albeit at a significantly higher computational cost.
This rich set of available tools, from Monte Carlo Dropout to Gaussian Processes, provides the technical foundation for building the next generation of uncertainty-aware systems. The critical challenge, which we address next, is how to integrate these methods into a coherent design and evaluation philosophy for modern SLAM.

\section{DISCUSSION AND FUTURE DIRECTIONS}

The arguments presented in this paper advocate for a fundamental shift in the design and evaluation philosophy of SLAM systems. This move towards uncertainty-awareness, framed not as a forgotten feature but as a central design challenge, has several profound implications for the future of the field.

\textit{From Uncertainty-Awareness to Consistency-Awareness.} Firstly, the community must move beyond simply using uncertainty to demanding its consistency. The standard output of a SLAM system should no longer be just $(\hat{\mathbf{x}}, \mathbf{P})$, but an estimate where $\mathbf{P}$ itself is verifiably consistent. This means developing metrics and methods that can detect when a system becomes overconfident, addressing the core problem of internal consistency failure \cite{castellanos2004limits}. Emerging frameworks like Gaussian Belief Propagation \cite{hug2024hyperion}, which treat uncertainty as a first-class citizen, are a step in this direction.

\textit{Scalable Covariance as a Grand Challenge.} Secondly, we must frame the recovery of uncertainty not as a post-processing step, but as a core algorithmic challenge. Instead of lamenting the computational cost, research efforts should focus on \textit{scalable covariance approximation} techniques. This includes methods for efficient marginalization and on-demand recovery, as demonstrated in active SLAM \cite{placed2022explorbslam},  while pushing these techniques towards the online performance required for robust autonomy, and exploring non-Gaussian uncertainty representations that can capture complex phenomena like multi-modality in ambiguous environments.

\textit{Bridging Probabilistic and Learning-based Methods.} Finally, as neural networks become integral to the perception pipeline, equipping them with robust uncertainty quantification mechanisms is critical. The next frontier will involve integrating these approaches with explainable AI (XAI) to move from uncertainty-awareness to \textit{uncertainty-understanding}, enabling robots that can diagnose and articulate the reasons for their potential failures in a human-interpretable manner \cite{qin2024towards}.

\section{CONCLUSIONS}




This paper has argued that the dominant SLAM evaluation paradigm, with its relentless focus on geometric accuracy on static benchmarks, has led the community into the \textit{SLAM Confidence Trap}. We have built a generation of systems that, while impressively accurate in controlled settings, are fundamentally brittle and lack the crucial capacity for real time self assessment. This trend, born from a historical shift away from consistent probabilistic modeling towards pure optimization, has produced algorithms that cannot robustly answer the most critical question for any autonomous agent: \enquote{How confident am I in my own estimate?}.


Escaping this trap requires a fundamental paradigm shift. Uncertainty estimation cannot remain a post processing afterthought; it must be treated as a core, indispensable component of any system designed for the real world. Therefore, we call for a collective re-evaluation of our metrics for success, moving beyond the simple pursuit of lower trajectory error. The future of SLAM research must be defined by the development of systems that are demonstrably robust and self aware. By making consistent uncertainty a first class citizen in our designs, we will pave the way for the next generation of autonomous systems that can be trusted to operate safely and effectively in our complex and unpredictable world.




\bibliographystyle{IEEEtran}
\bibliography{IEEEabrv,biblio}

\end{document}